\documentclass[]{imag-ms-template}

\usepackage{graphicx}
\usepackage{wrapfig}
\usepackage{amsmath}
\usepackage{amssymb}
\usepackage{booktabs}
\usepackage{enumitem}
\usepackage{gensymb}
\usepackage{algpseudocode}
\usepackage{xcolor}
\usepackage{subcaption}

\usepackage{dirtytalk}

\usepackage{multirow}
\usepackage{url}
\usepackage[symbol]{footmisc} 

\usepackage{float}

\title{OpenMAP-BrainAge: Generalizable and Interpretable Brain Age Predictor}
\date{} 

\author{Pengyu Kan,$^{1}$ Craig Jones,$^{1}$ Kenichi Oishi $^{2\dagger}$\\
for the Alzheimer's Disease Neuroimaging Initiative$^\ast$\\
and the Australian Imaging Biomarkers and Lifestyle flagship study of aging $^{\ast\ast}$\\
{\small $^{1}$Department of Computer Science, Johns Hopkins University, Baltimore, MD, USA}\\
{\small $^{2}$The Russell H. Morgan Department of Radiology and Radiological Science,}\\
{\small The Johns Hopkins University School of Medicine, Baltimore, MD, USA}\\
{\small $^\dagger$Correspondence:  koishi2@jhmi.edu}\\
}

\begin{document} 

\maketitle


\footnotetext{$^\ast$ Data used in preparation of this article were obtained from the Alzheimer's Disease Neuroimaging Initiative (ADNI) database (adni.loni.usc.edu). As such, the investigators within the ADNI contributed to the design and implementation of ADNI and/or provided data but did not participate in the analysis or writing of this report. A complete listing of ADNI investigators can be found at: \href{http://adni.loni.usc.edu/wp-content/uploads/how_to_apply/ADNI_Acknowledgement_List.pdf}{http://adni.loni.usc.edu/wp-content/uploads/how\_to\_apply/ADNI\_Acknowledgement\_List.pdf}} 

\footnotetext{$^{\ast\ast}$ Data used in the preparation of this article was obtained from the Australian Imaging Biomarkers and Lifestyle flagship study of ageing (AIBL) funded by the Commonwealth Scientific and Industrial Research Organisation (CSIRO),which was made available at the ADNI database (www.loni.usc.edu/ADNI). The AIBL researchers contributed data but did not participate in the analysis or writing of this report. AIBL researchers are listed at \href{www.aibl.csiro.au}{www.aibl.csiro.au}.}

\begin{abstract}

\textbf{Purpose}: 

To develop an age prediction model which is interpretable and robust to demographic and technological variances in brain MRI scans.

\textbf{Materials and Methods}: 

We propose a transformer-based architecture that leverages self-supervised pre-training on large-scale datasets. Our model processes pseudo-3D T1-weighted MRI scans from three anatomical views and incorporates brain volumetric information. By introducing a stem architecture, we reduce the conventional quadratic complexity of transformer models to linear complexity, enabling scalability for high-dimensional MRI data. We trained our model on ADNI2 $\&$ 3 (N=1348) and OASIS3 (N=716) datasets (age range: 42 - 95) from North America, with an 8:1:1 split for train, validation and test. Then, we validated it on the AIBL dataset (N=768, age range: 60 - 92) from Australia.

\textbf{Results}: 

We achieved an MAE of 3.65 years on ADNI2 $\&$ 3 and OASIS3 test set and a high generalizability of MAE of 3.54 years on AIBL. There was a notable increase in brain age gap (BAG) across cognitive groups, with mean of 0.15 years ($95\%$ CI: [-0.22, 0.51]) in CN, 2.55 years ([2.40, 2.70]) in MCI, 6.12 years ([5.82, 6.43]) in AD. Additionally,  significant negative correlation between BAG and cognitive scores was observed, with correlation coefficient of -0.185 (p $<$ 0.001) for MoCA and -0.231 (p $<$ 0.001) for MMSE. Gradient-based feature attribution highlighted ventricles and white matter structures as key regions influenced by brain aging.

\textbf{Conclusion}: 

Our model effectively fused information from different views and volumetric information to achieve state-of-the-art brain age prediction accuracy, improved generalizability and interpretability with association to neurodegenerative disorders.

\end{abstract}
\newpage
\section{Introduction}
\label{sec:intro}

The human brain experiences a complex structural development along the aging process, such as cortical thinning and ventricular enlargement \cite{giorgio2010age, uchida2024acceleration}. An individual's brain may age at a different rate, resulting in a deviation from their chronological age. The \textit{brain age gap} (BAG) quantifies the difference between biological and chronological ages \cite{he2021global} (Appendix Fig. \ref{fig:age_prediction}) and has been linked to neurodegenerative conditions such as Alzheimer’s disease (AD) and mild cognitive impairment (MCI) \cite{bashyam2020mri, siegel2024transformers, zhao2023modeling}. Therefore, being able to estimate the brain age gap may lead to the detection of neurodegenerative conditions.

BAG estimators have utilized convolutional neural networks (CNNs) \cite{lecun1989backpropagation} to extract spatial features from brain MRIs \cite{bashyam2020mri, lee2022deep, peng2021accurate}. While CNNs effectively capture global patterns via sliding kernels, they may overlook fine-grained local details. Recent transformer-based architectures \cite{he2021global, siegel2024transformers, zhao2023modeling} have been explored for their ability to capture long-range dependencies and adaptively weigh attention across spatial locations and inputs \cite{vaswani2017attention}. Though some models \cite{he2021global} reported errors below 3 years when trained on large-scale multi-cohort datasets, variations in datasets and protocols hinder direct performance comparisons. Age bias in training data can skew estimations towards the group mean \cite{bashyam2020mri, peng2021accurate}, and models often show limited transferability across diverse demographics and imaging protocols. The quadratic computational cost of transformers also limits scalability for full 3D MRI inputs \cite{vaswani2017attention}.

To mitigate these challenges, we introduce OpenMAP-BrainAge. Since pretrained foundation models enhance generalization across unseen domains \cite{bao2021beit, wang2024scaling}, we utilized a transformer-based foundation model \cite{wang2024scaling}, which was pretrained on 52 robotic proprioception and vision datasets in the real world and collected across diverse environments and hardware configurations, to address the bias caused by variability in MRI scanning techniques. Since incorporating local information yields better brain age predictions \cite{he2021global, zhao2023modeling}, we propose pseudo-3D inputs, which sample central chunks from 2D sagittal, coronal and axial slices, thereby balancing local details and global contexts, while mitigating model complexity. Additionally, volumes from 280 brain structures \cite{nishimaki2024openmap} were included, as their changes are associated with aging \cite{zhao2023modeling}. Our architecture fuses these multiview and local volumetric data through a modified self-attention mechanism inspired by the stem-and-trunk design in \cite{wang2024scaling}. The model was validated on the Alzheimer’s Disease Neuroimaging Initiative 2 \& 3 (ADNI 2\&3) \cite{petersen2010alzheimer} and Open Access Series of Imaging Studies 3 (OASIS 3) \cite{lamontagne2019oasis} datasets, and its robustness was further assessed across domains and cohorts using the Australian Imaging Biomarkers and Lifestyle Study of Ageing (AIBL) dataset \cite{ellis2009australian}.

\section{Materials and Methods}
\label{sec:method}

\subsection{OpenMAP-BrainAge}
 
\begin{figure}[H]
    \centering
    \captionsetup[subfigure]{font=footnotesize}
   \begin{subfigure}[b]{0.8\textwidth}
        \includegraphics[width=\textwidth]{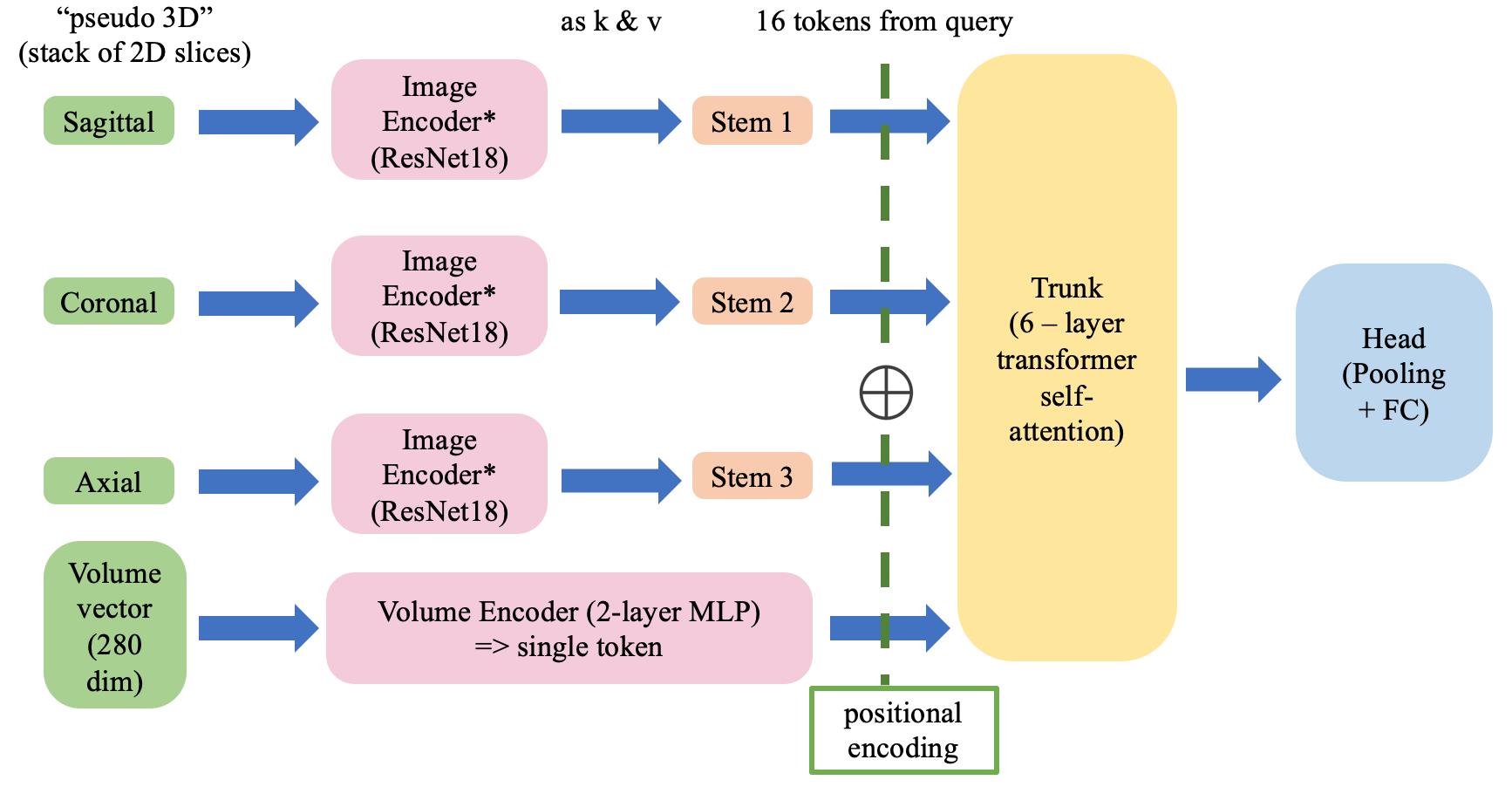}
        \caption{Illustration of OpenMAP-BrainAge. The proposed model takes "pseudo" 3D inputs, composed of central chunks from three anatomical views, and volumetric vectors from 280 brain regions. The shared Image Encoder extracts image features that are processed through distinct stems. The Trunk fuses the embedding from different views and volumetric information. The fused features are decided through the Head.}
      \label{fig:multiview-transformer}
    \end{subfigure}
    \hfill
    \begin{subfigure}[b]{0.8\textwidth}
        \includegraphics[width=\textwidth]{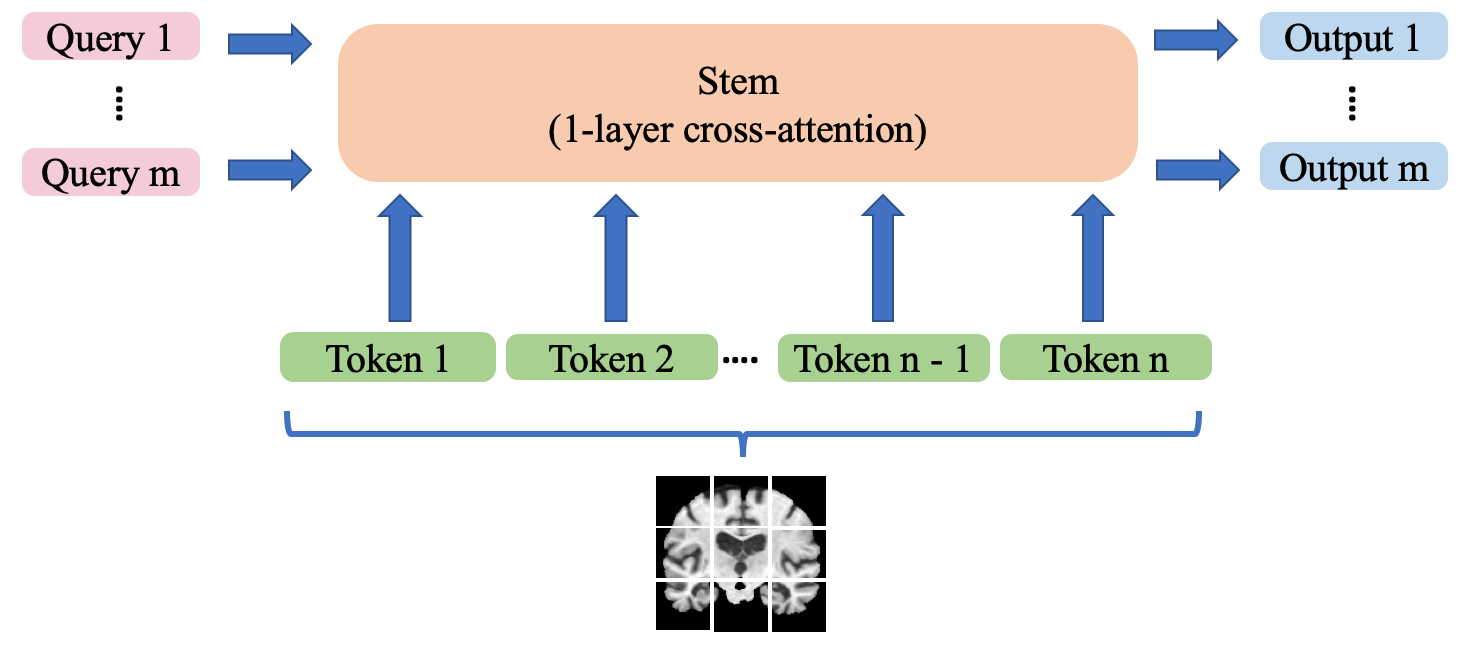}
        \caption{Stem architecture. We introduce a fixed number $m$ of learnable queries (much fewer than the number $n$ of tokens from the input), as following L. Wang et al. \cite{wang2024scaling}. Then, we capture the relationship between these learnable queries and the keys from the input to generate embeddings for the input.}
         \label{fig:stem}
    \end{subfigure}
    \caption{Architecture design of OpenMAP-BrainAge.}
    \label{fig:architecture_design}
\end{figure}

 OpenMAP-BrainAge is a transformer-based model designed to predict brain age from MRI and volumetric data (Fig. \ref{fig:multiview-transformer}). The architecture applies attention mechanisms (Appendix \ref{ap:attn}) to dynamically fuse information across multiple anatomical views and data types. The model consists of four core components: encoders, stems, a trunk, and task-specific heads. The encoders project raw inputs into embeddings for later processing and fusing. For MRI data, we use a 3D ResNet18 \cite{hara3dcnns} initialized with weights pretrained on the Kinetics-700 video dataset \cite{carreira2019short} to capture the spatial features. For volumetric data, a lightweight two-layer multilayer perceptron (MLP) maps numeric inputs into the same feature space. The stem module \cite{wang2024scaling} (Fig. \ref{fig:stem}) compresses the image encoders' tokenized outputs into a fixed-size representation using learnable queries and attention. This reduces computational cost and enables efficient fusion across inputs in linear time and space complexity (Appendix \ref{ap:complexity_stem}). The trunk is a transformer block that integrates features from stem outputs for each anatomical view and from the 280-region volumes, using self-attention \cite{vaswani2017attention} to capture relations across views and data types. We adapted the pre-trained trunk from various tasks across 52 robotic datasets \cite{wang2024scaling}. Finally, the head projects the trunk output to the prediction task, where we used a regression MLP for estimating biological age. This design allows flexible adaptation to multiple data types while maintaining efficiency through shared modules and pretrained initialization.

\subsection{Datasets}
We trained and validated our model using the cognitive normal (CN) participants from the ADNI 2\& 3 \cite{petersen2010alzheimer} and the OASIS 3 \cite{lamontagne2019oasis} datasets, which included participants aged 42 to 95 from North America. It was assumed the CN group had aligned biological and chronological ages. Participants were randomly divided into $8:1:1$ for train, validation, and test purposes (4630 scans, 584 scans, and 543 scans in each), ensuring no subjects overlap across the sets. Additionally, for correlation analyses with clinical variables, we studied 1831 scans for AD and 3872 scans for MCI.

We preprocessed the T1-weighted MRI scans using the OpenMAP-T1 \cite{nishimaki2024openmap}, which skull-stripped the brain and rigidly aligned it to MNI space. Each scan was downsampled to $2\times2\times2$ mm resolution to reduce input size. To construct the pseudo-3D inputs, we cropped $128\times 128\times 30$ voxel volumes from the center of each anatomical view: sagittal, coronal, and axial. Then, we applied the min-max normalization on the intensity. The data augmentation during training consisted of random rotations ($\pm 20^{\degree}$) and random transformations ($\pm 20$ pixels), each at $p=50\%$.
\subsection{External Cohort}
To evaluate the robustness and generalizability, we estimated the brain age on the CN group from the AIBL dataset \cite{ellis2009australian}, which contained 648 subjects (60 to 92 years old). AIBL is an Australian dataset that differs geographically from the training data, although it follows similar scanning protocols and preprocessing steps as the ADNI dataset.

\subsection{Training Setup}
The BAG training was formulated as a regression task, where the mean squared error (MSE) between predicted ages $\hat{\mathbf{y}}\in\mathbb{R}^N$ and the chronological ages (ground truth labels) $\mathbf{y}\in\mathbb{R}^N$ for $N$ subjects was quantified, as defined in Eq. \ref{eq:training_loss}.
\begin{align}
    L_\text{training} = \text{MSE }(\hat{\mathbf{y}}, \mathbf{y})=\frac{1}{N}\|\hat{\mathbf{y}} - \mathbf{y}\|_2^2 = \frac{1}{N}\sum\limits_{i=1}^N |\hat{y}_i-y_i|^2
    \label{eq:training_loss}
\end{align}

We trained with the Adam Optimizer \cite{kingma2014adam} with an initial learning rate of $3\times 10^{-5}$, decaying by a factor of $0.1$ for every 10 epochs; batch size of 32; and 200 epochs. 

\subsection{Performance Evaluation}

The model accuracy was assessed with the mean absolute error (MAE) on the test set, as shown in Eq. \ref{eq:valid_loss}.
\begin{align}
L_\text{valid} = \text{MAE }(\hat{\mathbf{y}}, \mathbf{y})=\frac{1}{N}\|\hat{\mathbf{y}} - \mathbf{y}\|_1^1 = \frac{1}{N}\sum\limits_{i=1}^N |\hat{y}_i-y_i|
\label{eq:valid_loss}  
\end{align}
For clinical analyses, we computed the BAG value and defined as:
\begin{equation}
    \text{BAG} = \hat{y}_i-y_i
\end{equation}

\subsection{Comparison with State-of-the-Art Neural Networks}
\label{sec:exp_nn_comp}

To compare our model with previous published work, we retrained the following state-of-the-art neural networks under our setup: (1) multilayer perceptron (MLP): a two-layer fully connected network with ReLU activation to run the regression task. (2) 3D ResNet18 \cite{hara3dcnns, he2016deep}: pretrained with a classification task over the Kinetic-700 video dataset \cite{carreira2019short}. We changed the final fully connected layer to fit the brain age prediction task and fine-tuned all layers with our training data. (3) Vision Transformer (ViT) \cite{dosovitskiy2020image}: we selected the ViT-Base model (12 layers, 768 hidden size, and 12 attention heads) pretrained through BERT self-supervised image pretraining process (BEiT) \cite{bao2021beit} over the ImageNet-1K dataset \cite{russakovsky2015imagenet}. The 2D convolutional patch embedding layer was replaced with a 3D version to handle MRI volumes. We fine-tuned all parameters over our training setup. (4) Simple Fully Convolutional Network (SFCN) \cite{peng2021accurate}: contained seven convolutional layers, batch normalization, ReLU activation and max pooling. We re-initialized the model with Kaiming uniform initialization \cite{he2015delving} and trained over our training dataset. (5) Global - Local Transformer \cite{he2021global}: contained VGG8 as the image feature backbone to extract information from both the global and local pathways and attention layer to merge the information. The Global - Local Transformer took 2D axial slices as input and averaged across these axial slices for the 3D scenario.

\section{Results}
\label{sec:results}

\subsection{Performance Comparison over CN Test}

Table 
compared the performance with the state-of-the-art neural networks as described in Sec. \ref{sec:exp_nn_comp} on four types of input: (1) Volume: the volumetric vector of 280-region parcellated by OpenMAP-T1 \cite{nishimaki2024openmap}; (2) Pseudo 3D - Coronal: the central chunks from coronal view only; (3) Pseudo 3D - three-view + Volume: combination of the central chunks from all three anatomical views and the volumetric vector; and (4) Full 3D: the whole 3D MRI scans. 

We observed that, using volumetric information only, MLP model yielded a high MAE of 11.43 years, indicating the volumetric information alone was insufficient for accurate brain age estimation. 3D ResNet18 model \cite{hara3dcnns, he2016deep} achieved its best performance with full 3D input (MAE of 3.61 years and R-value of 0.82), and slightly lower performance using central coronal slices (MAE of 3.78 years and R-value of 0.81). Surprisingly, it performed worst with the multiview and volumetric inputs, suggesting limited ability of 3D ResNet18 to fuse heterogeneous information. For 3D ViT \cite{dosovitskiy2020image}, due to its quadratic time and space complexity, full 3D input was not applicable with exceeding the limit of GPU capacity and it reached MAE of 3.63 years and R-value of 0.81 with taking central coronal slices. We re-trained SFCN \cite{peng2021accurate} and the Global-Local Transformer \cite{he2021global} over our training split using their original input format. SFCN and Global-Local Transformer achieved MAE of 5.94 and 3.87 years respectively, which were higher than the published literature results, potentially due to reduced training samples. In comparison, our proposed model achieved a competitive MAE of 3.65 years and R-value of 0.82, with better information fusion than 3D ResNet18 and notably less computation than transformer-based models (1.33 GFlops vs. 65.43 GFlops for 3D ViT and 1793.48 GFlops for Global-Local Transformer).

\begin{figure}[H]
    \centering
    \captionsetup[subfigure]{font=footnotesize}
   \begin{subfigure}[b]{0.3\textwidth}
        \includegraphics[width=\textwidth]{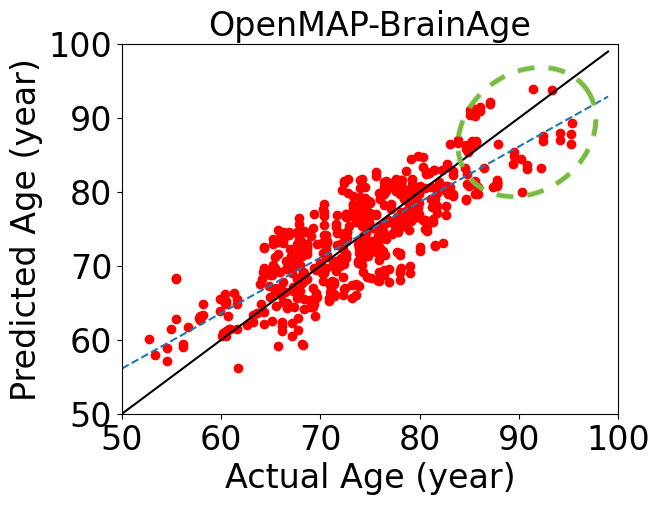}
        \captionsetup{justification=centering}
        \caption{
MAE: 3.65, std: 2.43, \\ R-value: 0.82.\\ 
 Pseudo 3D - three view + Volume.}
        \label{fig:CN_comp_a}
    \end{subfigure}
    \begin{subfigure}[b]{0.3\textwidth}
        \includegraphics[width=\textwidth]{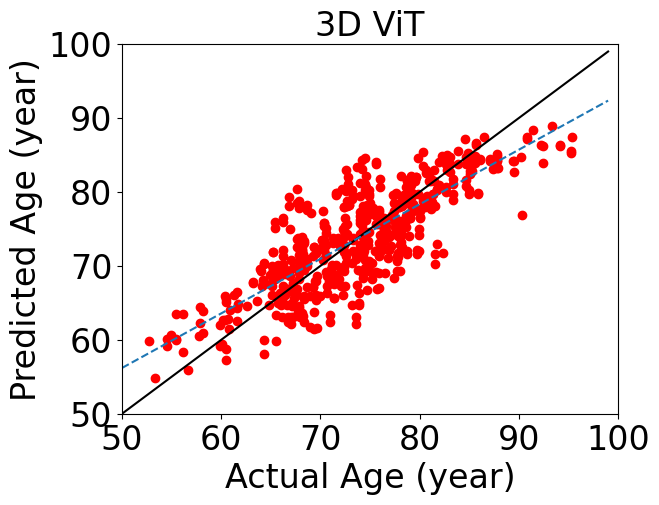}
         \captionsetup{justification=centering}
        \caption{
  MAE: 3.63, std: 2.79,\\   R-value: 0.81.\\   Pseudo 3D - Coronal.}
        \label{fig:CN_comp_b}
    \end{subfigure}
    \begin{subfigure}[b]{0.3\textwidth}
        \includegraphics[width=\textwidth]{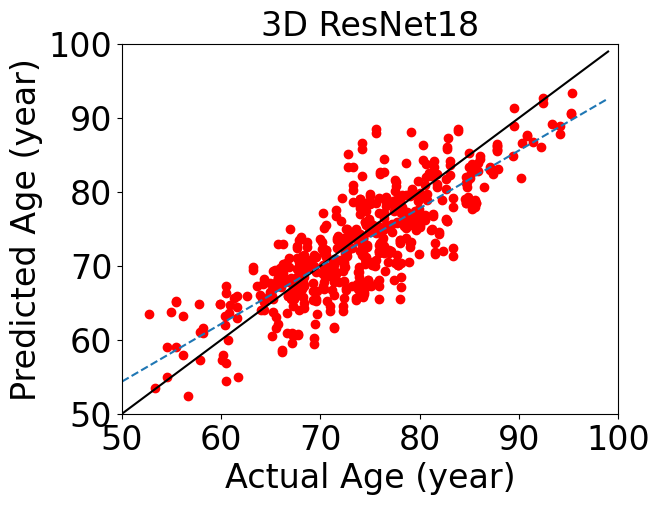}
         \captionsetup{justification=centering}
        \caption{
  MAE: 3.61, std: 2.70,\\ R-value: 0.82.\\  Full 3D.}
        \label{fig:CN_comp_c}
    \end{subfigure}
    \caption{Scatter of predictions versus actual ages on CN group of the test split of ADNI 2\&3 \cite{petersen2010alzheimer} and OASIS 3  \cite{lamontagne2019oasis}. (a) OpenMAP-BrainAge with ``Pseudo 3D - three view + Volume" input. (b) 3D ViT \cite{dosovitskiy2020image} with ``Pseudo 3D - Coronal" input. (c) 3D ResNet18 \cite{hara3dcnns, he2016deep} with ``Full 3D" input.  OpenMAP-BrainAge achieves a mostly evenly scatter of predictions (green circled region) on the elder age group (age over 85 years old), meanwhile other two models tend to under-estimate on elder group though reaching a similar mean absolute error (MAE) in general. Further, OpenMAP-BrainAge has the lowest standard deviation of error on predictions, which implies a higher confidence in the estimation.}
    \label{fig:CN_comp}
\end{figure}

In Fig. \ref{fig:CN_comp}, we selected three models with the best performance in Table \ref{tab:result_cn} to make a further comparison. Both 3D ViT and 3D ResNet18 tended to under-estimate the elder age group (age over 85 years old), which is a common issue \cite{peng2021accurate}. Furthermore, our proposed model had the lowest standard deviation of $2.43$ years than the others ($2.79$ years for 3D ViT and $2.7$ years for 3D ResNet18), indicating more consistent confident predictions in the estimation.

\subsection{Performance Comparison over External Cohort}
\label{subsec:result_external_corhort}

To test on domain shift robustness caused by geographical differences of subjects, we evaluated the performance in the CN group in the AIBL dataset \cite{ellis2009australian}. Table 
showed that our model maintained a consistent performance with MAE of 3.54 years as compared to 3.87 years originally (statistically insignificant difference with T-statistics 0.770 and P-value 0.442), despite a slight drop of R-value from 0.82 to 0.76. In contrast, most state-of-the-art models exhibited performance degradation with MAE increase of 0.14 to 0.61 years and R-value reduction of 0.1. These results demonstrated that our model is robust to domain shifts introduced by demographic variability.

\subsection{Analysis on Brain Disorders}
\label{subsec:result_disorder}

\begin{figure}[H]
    \centering
    \captionsetup[subfigure]{font=footnotesize}
    \begin{subfigure}[b]{0.3\textwidth}
        \includegraphics[width=\textwidth]{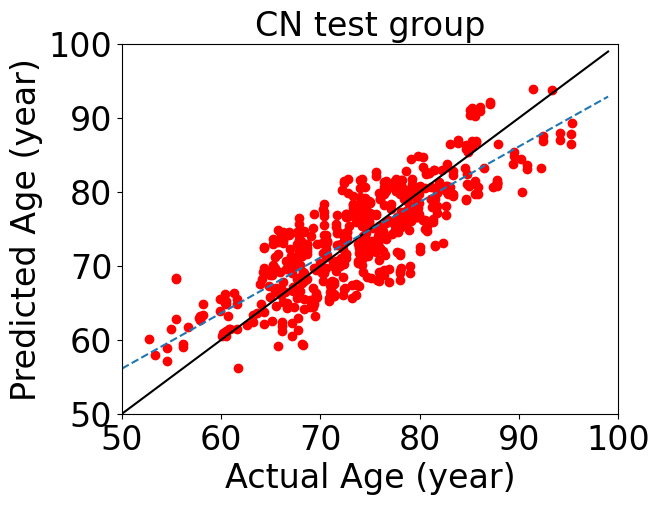}
         \captionsetup{justification=centering}
        \caption{MAE: 3.65, std: 2.43,\\  R-value: 0.82.}
        \label{fig:disorder_a}
    \end{subfigure}
    \begin{subfigure}[b]{0.3\textwidth}
        \includegraphics[width=\textwidth]{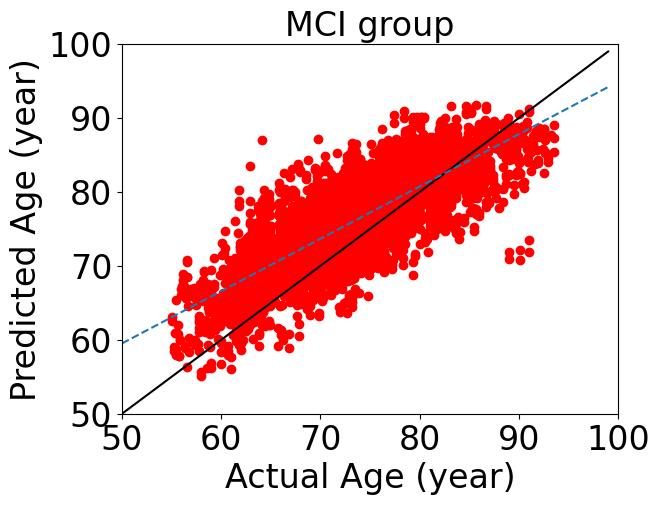}
         \captionsetup{justification=centering}
        \caption{MAE: 4.33, std: 3.2,\\ R-value: 0.71.}
        \label{fig:disorder_b}
    \end{subfigure}
    \begin{subfigure}[b]{0.3\textwidth}
        \includegraphics[width=\textwidth]{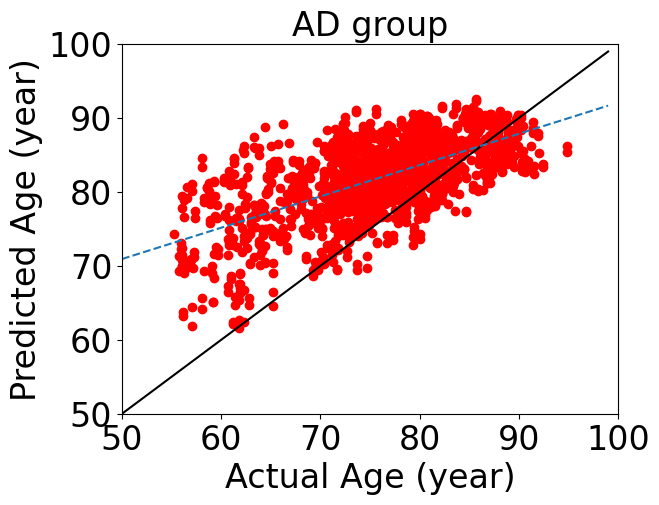}
         \captionsetup{justification=centering}
        \caption{MAE: 7.05, std: 5.16,\\  R-value: 0.42.}
        \label{fig:disorder_c}
    \end{subfigure}
    \hfill
    \begin{subfigure}[b]{0.28\textwidth}
        \includegraphics[width=\textwidth]{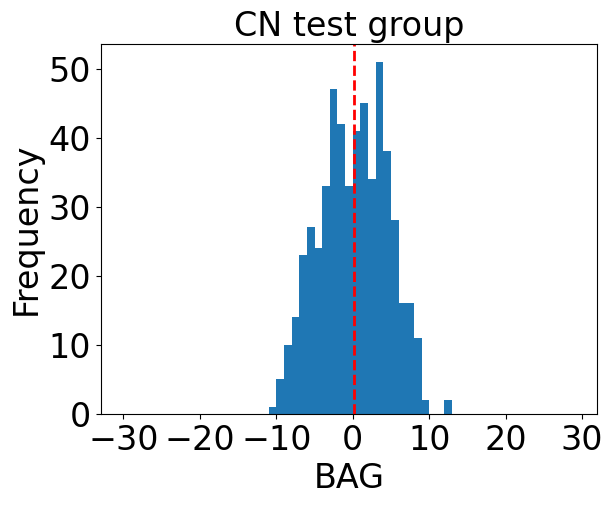}
         \captionsetup{justification=centering}
        \caption{Mean: 0.15.\\
        $95\%$ CI: (-0.22, 0.51).}
        \label{fig:disorder_d}
    \end{subfigure}
    \begin{subfigure}[b]{0.28\textwidth}
        \includegraphics[width=\textwidth]{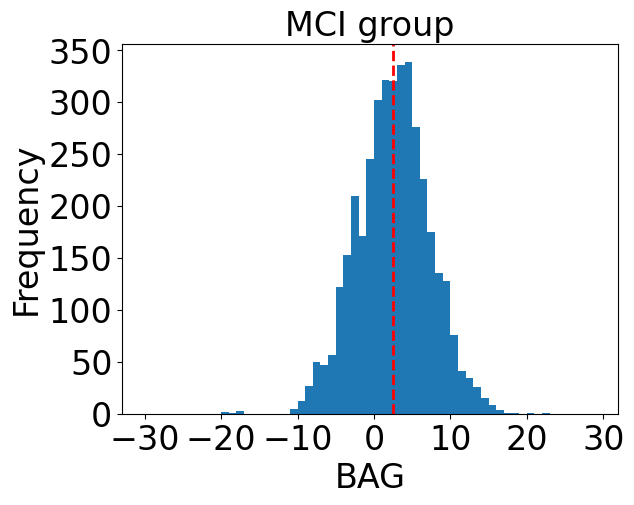}
         \captionsetup{justification=centering}
        \caption{Mean: 2.55.\\
        $95\%$ CI: (2.4, 2.7).}
        \label{fig:disorder_e}
    \end{subfigure}
    \begin{subfigure}[b]{0.28\textwidth}
        \includegraphics[width=\textwidth]{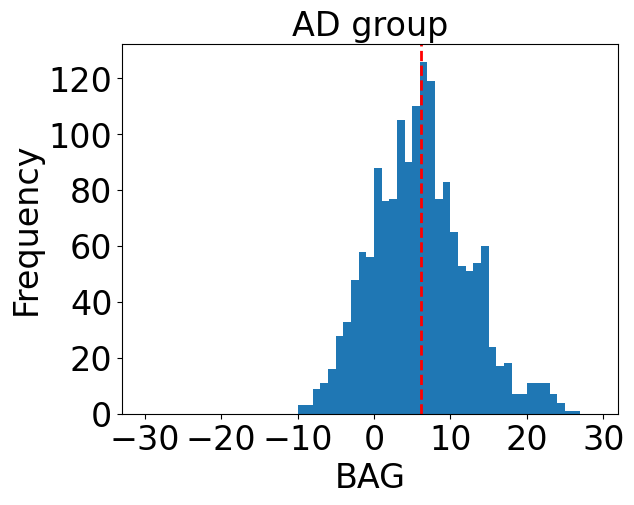}
         \captionsetup{justification=centering}
        \caption{Mean: 6.12.\\
        $95\%$ CI: (5.82, 6.43).}
        \label{fig:disorder_f}
    \end{subfigure}
    \caption{Analysis of brain aging across different subject groups. In the first row, we plotted the prediction versus the actual age of each scan. In the second row, we plotted the histogram of the brain age gap (BAG). The predictions get greater overestimation and the mean BAG values are increasing, as the subject groups suffering more severe disorders.}
    \label{fig:disorder}
\end{figure}

We analyzed the BAG across CN, MCI and AD patients. In Fig. \ref{fig:disorder}, for the CN group, the predicted age was highly correlated with the actual age ($R=0.82$). The BAG values were centered at mean of $0.15$ years, with $95\%$ Confidence Interval (CI) of $(-0.22, 0.51)$. For the MCI group, we observed a trend of overestimation, with BAG values centered at a mean of 2.55 years and a $95\%$ CI of $(2.40, 2.70)$. For the AD group, representing the most severe brain disorder among the three, ages of the majority of scans were overestimated. The BAG values notably increased, with a mean of 6.12 years and a $95\%$ CI of 
$(5.82, 6.43)$. Thus, MCI and AD appeared to advance the brain aging process.

\begin{figure}[H]
    \centering
    \captionsetup[subfigure]{font=footnotesize}
    \begin{subfigure}[b]{0.32\textwidth}
        \includegraphics[width=\textwidth]{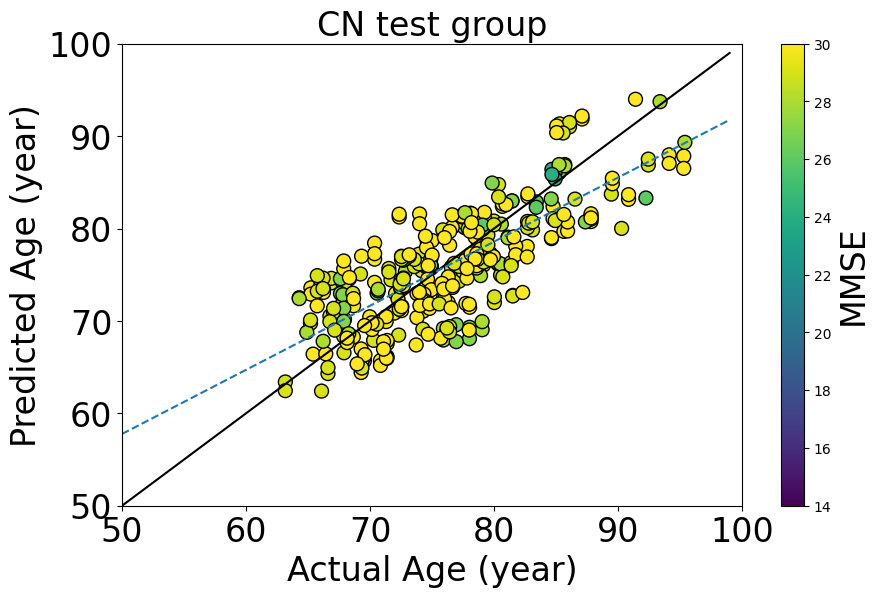}
        \caption{}
        \label{fig:mmse_analysis_a}
    \end{subfigure}
    \begin{subfigure}[b]{0.32\textwidth}
        \includegraphics[width=\textwidth]{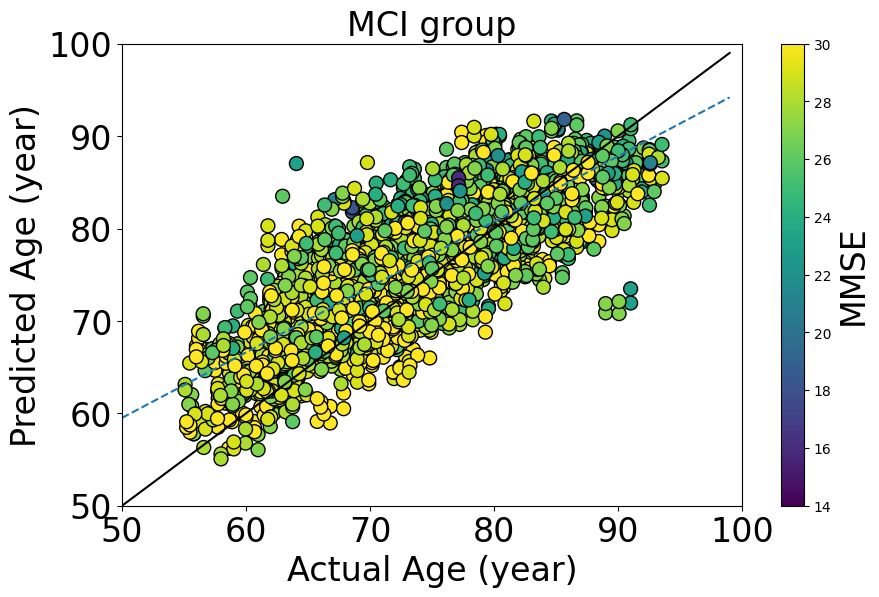}
        \caption{}
        \label{fig:mmse_analysis_b}
    \end{subfigure}
    \begin{subfigure}[b]{0.32\textwidth}
        \includegraphics[width=\textwidth]{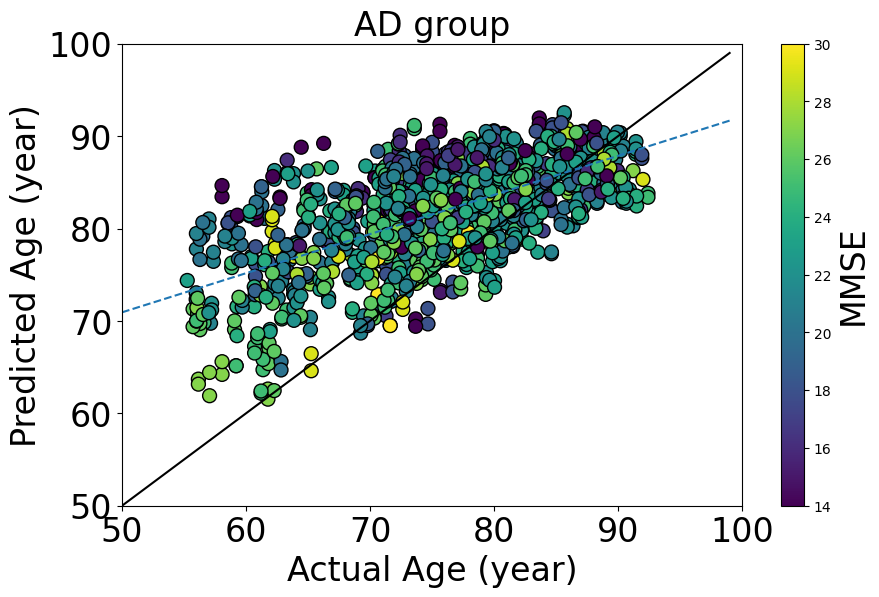}
        \caption{}
        \label{fig:mmse_analysis_c}
    \end{subfigure}
    \newline
    \begin{subfigure}[b]{0.32\textwidth}
        \includegraphics[width=\textwidth]{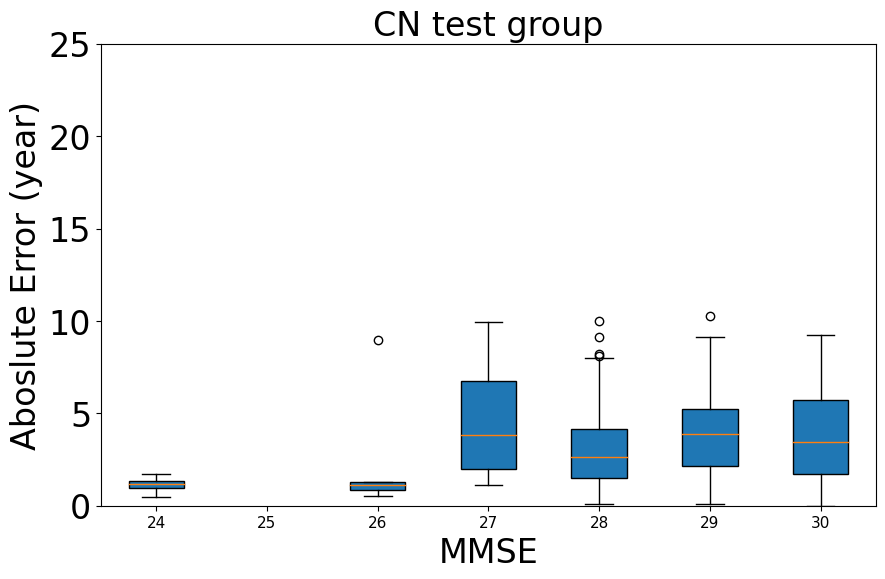}
         \captionsetup{justification=centering}
        \caption{R-value: $-0.0405$\\ P-value: $0.4560$}
        \label{fig:mmse_analysis_d}
    \end{subfigure}
    \begin{subfigure}[b]{0.32\textwidth}
        \includegraphics[width=\textwidth]{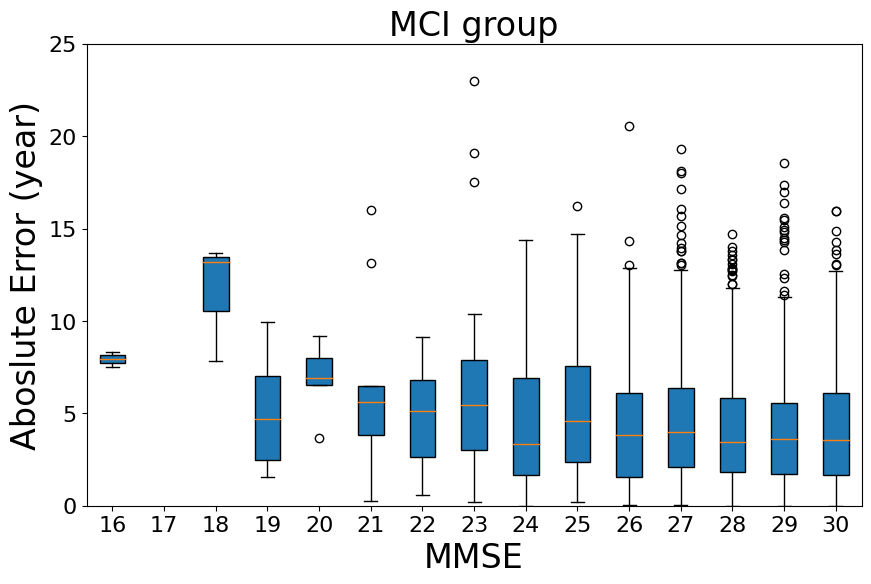}
         \captionsetup{justification=centering}
        \caption{R-value: $-0.0934$\\ P-value $< 0.001$}
        \label{fig:mmse_analysis_e}
    \end{subfigure}
     \begin{subfigure}[b]{0.32\textwidth}
        \includegraphics[width=\textwidth]{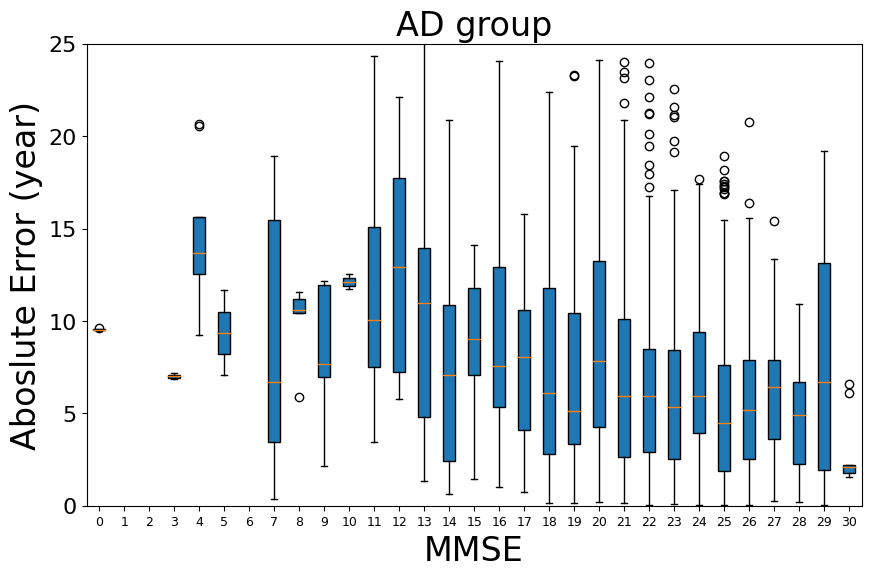}
         \captionsetup{justification=centering}
        \caption{R-value: $-0.1854$\\ P-value $< 0.001$}
        \label{fig:mmse_analysis_f}
    \end{subfigure}
    \caption{Correlation analysis between brain aging and MMSE.}
    \label{fig:mmse_analysis}
\end{figure}

\begin{figure}[H]
\captionsetup[subfigure]{font=footnotesize}
    \begin{subfigure}[b]{0.32\textwidth}
        \includegraphics[width=\textwidth]{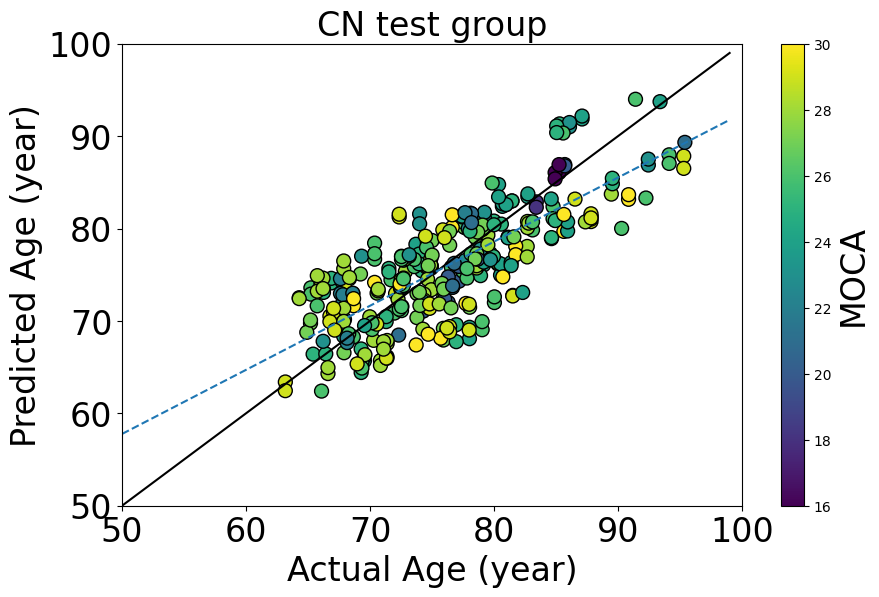}
        \caption{}
        \label{fig:moca_analysis_a}
    \end{subfigure}
     \begin{subfigure}[b]{0.32\textwidth}
        \includegraphics[width=\textwidth]{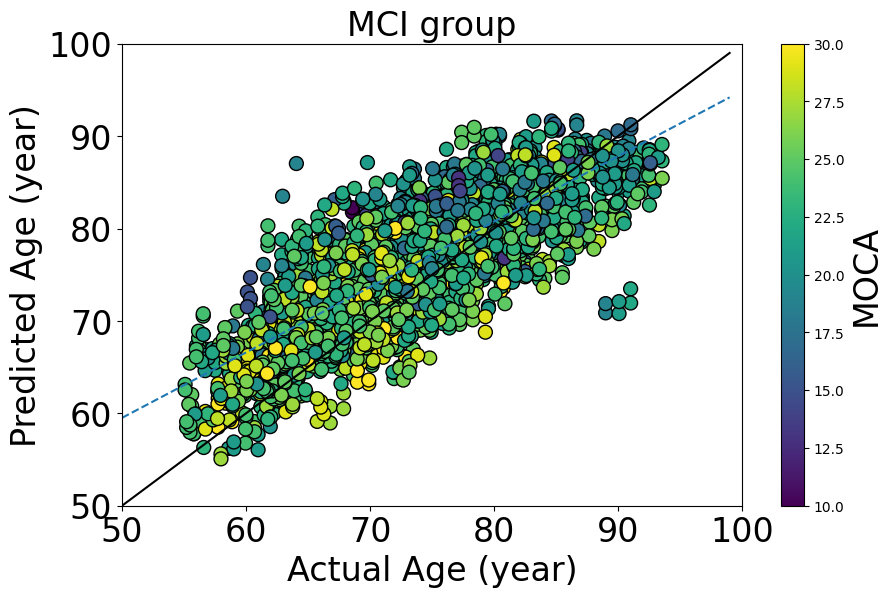}
        \caption{}
        \label{fig:moca_analysis_b}
    \end{subfigure}
     \begin{subfigure}[b]{0.32\textwidth}
        \includegraphics[width=\textwidth]{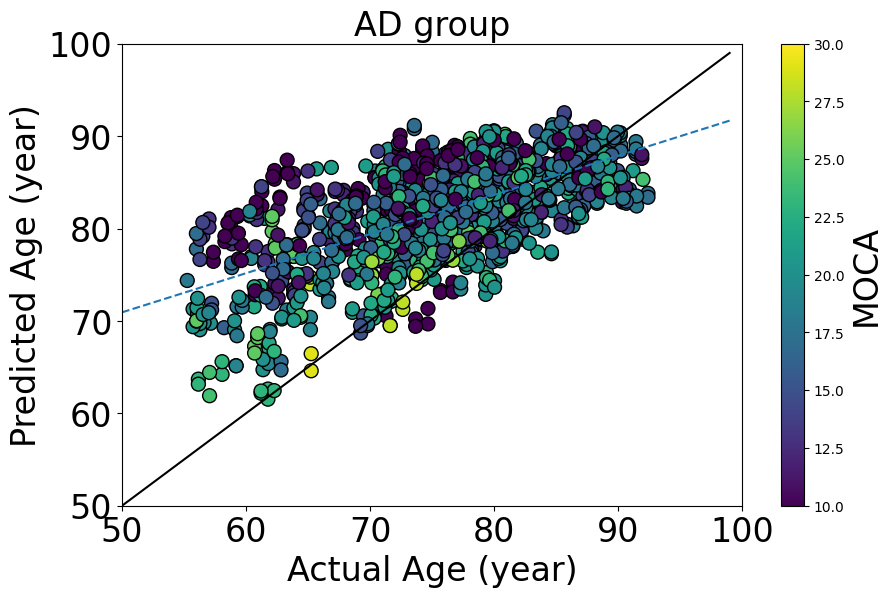}
        \caption{}
        \label{fig:moca_analysis_c}
    \end{subfigure}
    \newline
    \begin{subfigure}[b]{0.32\textwidth}
        \includegraphics[width=\textwidth]{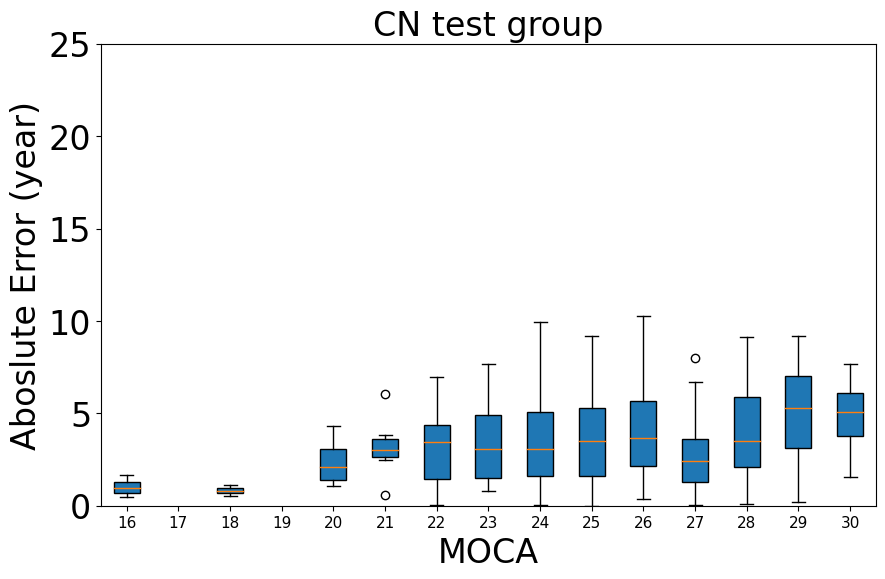}
         \captionsetup{justification=centering}
        \caption{R-value: $-0.1192$\\ P-value: 0.0283}
        \label{fig:moca_analysis_d}
    \end{subfigure}
     \begin{subfigure}[b]{0.32\textwidth}
        \includegraphics[width=\textwidth]{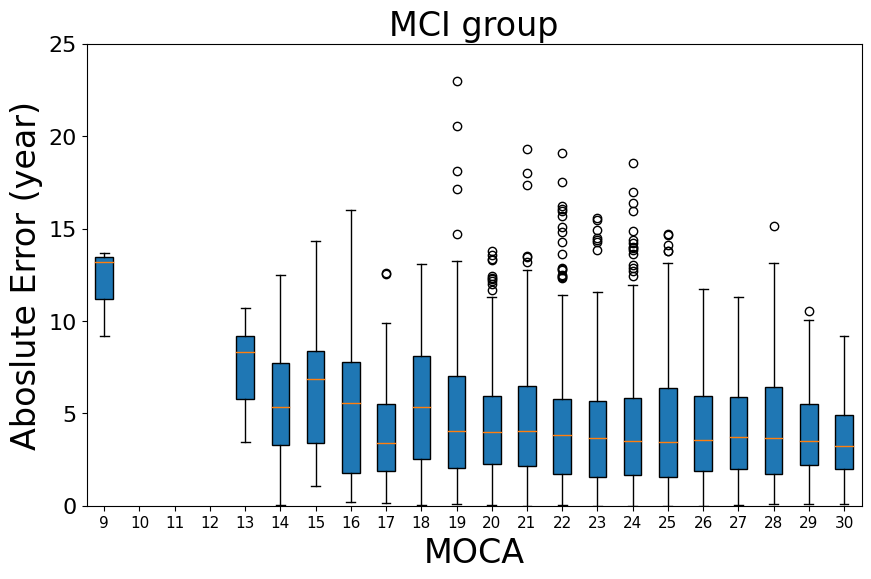}
         \captionsetup{justification=centering}
        \caption{R-value: $-0.1239$\\ P-value $< 0.001$}
        \label{fig:moca_analysis_e}
    \end{subfigure}
    \begin{subfigure}[b]{0.32\textwidth}
        \includegraphics[width=\textwidth]{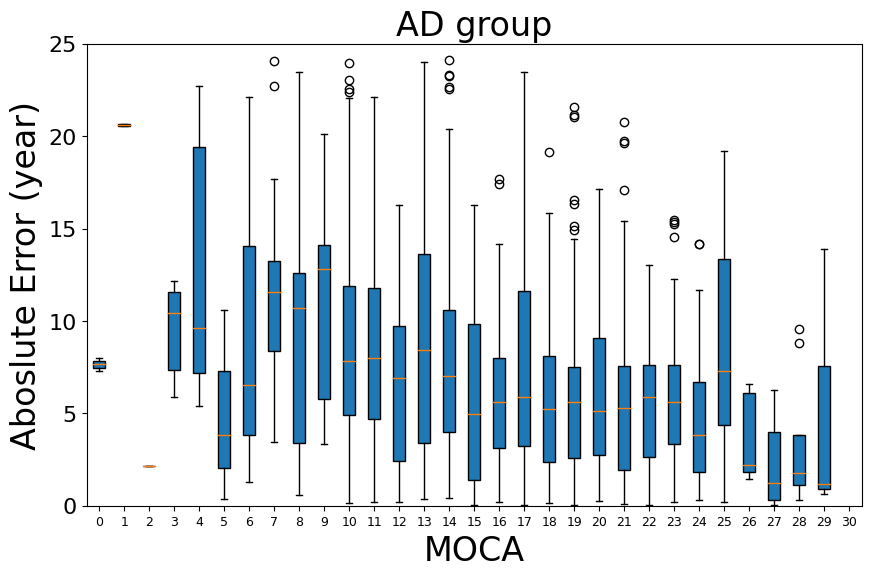}
         \captionsetup{justification=centering}
        \caption{R-value: $-0.2311$\\ P-value $< 0.001$}
        \label{fig:moca_analysis_f}
    \end{subfigure}
    \caption{Correlation analysis between brain aging and MoCA.}
    \label{fig:moca_analysis}
\end{figure}

We further analyzed the association between brain aging and cognitive scores, including MMSE (Fig. \ref{fig:mmse_analysis}) and MoCA (Fig. \ref{fig:moca_analysis}). Scans with lower cognitive scores, which indicate a more severe brain disorder, tended to increase the predicted age. In the CN group, the absolute error between the prediction and actual age was not significantly associated with the cognitive scores, with P-value $0.4560$ in MMSE and P-value $0.0283$ in MoCA. In contrast, in MCI and AD groups, we observed a significant negative correlation between the absolute error and the cognitive scores, with R-value of $-0.0934$ (P-value $< 0.001$) and R-value of $-0.1854$ (P-value $< 0.001$) in MMSE and R-value of $-0.1239$ (P-value $< 0.001$) and R-value of $-0.2311$ (P-value $< 0.001$) in MoCA respectively.

\subsection{Interpretation with Gradient Mapping}
\label{subsec:result_grad_map}

The average gradient maps (Appendix \ref{ap:grad_map}) were computed across the test split to identify the brain regions most influential in age prediction. To quantitatively compare how each region acts in the brain aging process, we normalized the gradient by the volume of each region. Appendix Fig. \ref{fig:region_ranking} ranked and listed the top 15 critical regions that determine the model's prediction, which implied that these 15 regions are highly influenced by the brain aging process. 

\begin{figure}[H]
    \captionsetup[subfigure]{font=footnotesize}
    \centering
    \begin{subfigure}[b]{0.23\textwidth}
        \includegraphics[width=1\textwidth]{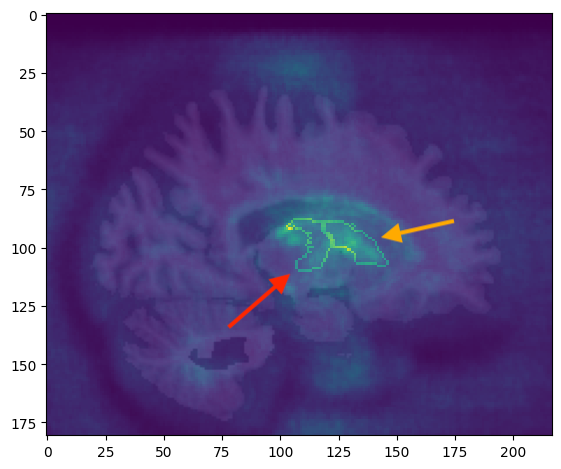}
        \caption{PLIC-R and ALIC-R.}
        \label{fig:activation_a}
    \end{subfigure}
    \begin{subfigure}[b]{0.23\textwidth}
        \includegraphics[width=\textwidth]{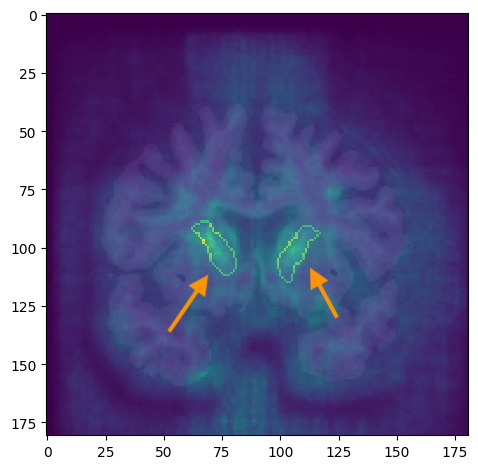}
        \caption{AILC-R and ALIC-L.}
        \label{fig:activation_b}
    \end{subfigure}
    \begin{subfigure}[b]{0.23\textwidth}
        \includegraphics[width=\textwidth]{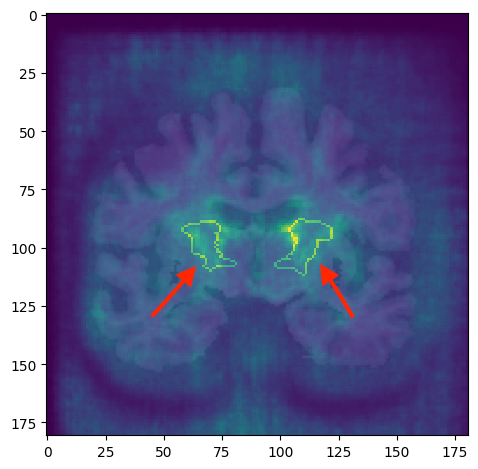}
        \caption{PLIC-R and PLIC-L.}
        \label{fig:activation_c}
    \end{subfigure}
     \begin{subfigure}[b]{0.23\textwidth}
        \includegraphics[width=\textwidth]{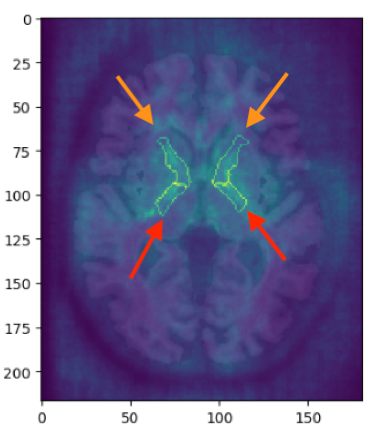}
        \caption{ALIC and PLIC.}
        \label{fig:activation_d}
    \end{subfigure}
    \hfill
    \begin{subfigure}[b]{0.23\textwidth}
        \includegraphics[width=\textwidth]{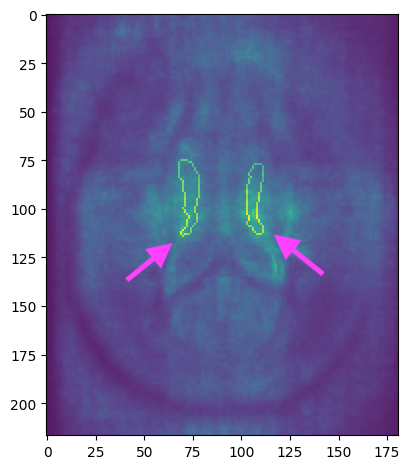}
        \caption{Caudate-L and -R.}
        \label{fig:activation_e}
    \end{subfigure}
    \begin{subfigure}[b]{0.23\textwidth}
        \includegraphics[width=\textwidth]{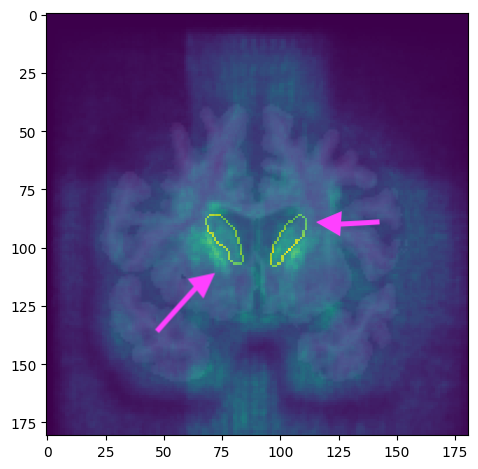}
        \caption{Caudate-L and -R.}
        \label{fig:activation_f}
    \end{subfigure}
    \begin{subfigure}[b]{0.23\textwidth}
        \includegraphics[width=\textwidth]{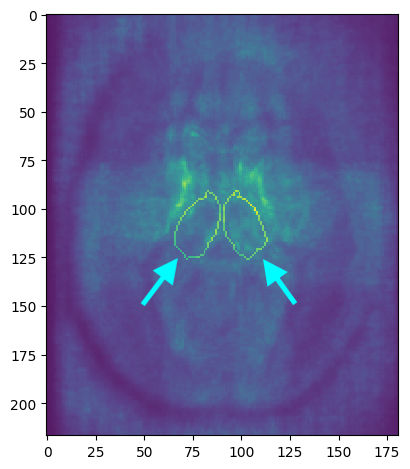}
        \caption{Thalamus-L and -R.}
        \label{fig:activation_g}
    \end{subfigure}
    \begin{subfigure}[b]{0.23\textwidth}
        \includegraphics[width=\textwidth]{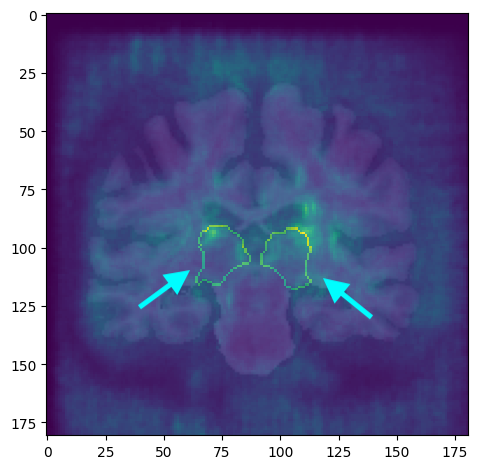}
        \caption{Thalamus-L and -R.}
        \label{fig:activation_h}
    \end{subfigure}
   
    \caption{Average gradient mapping of input with respect to training loss across the test split. Anterior Limb of Internal Capsule (ALIC; orange) and Posterior Limb of Internal Capsule (PLIC; red) are the top critical regions, determining our proposed model's estimation (Fig. \ref{fig:activation_a} - \ref{fig:activation_d}). Caudate nuclei (pink) and thalami (blue) are also highlighted as top 15 critical regions (Fig. \ref{fig:activation_e} - \ref{fig:activation_h}).}
    \label{fig:activation}
\end{figure}

The majority of the top-ranked regions were white matter structures and ventricles, aligning with the findings in the previous studies \cite{giorgio2010age, uchida2024acceleration}. Notably, Anterior Limb of Internal Capsule (ALIC) and Posterior Limb of Internal Capsule (PLIC) were listed as the top four most critical regions. Their volumes showed significant negative correlations with age, with ALIC exhibiting an R-value of $-0.21$ (P-value $<0.001$) and PLIC an R-value of $-0.27$ (P-value $<0.001$) (Appendix Fig. \ref{fig:vol_age}). In addition, several gray matter regions were also shown to be influenced by the aging process with decreasing volume, including the Caudate Nucleus and the Thalamus \cite{bauer2015significance, fama2015thalamic}. In Fig. \ref{fig:activation}, we visualized the gradient map and labeled these regions.

\section{Discussion}
\label{sec:discussion}

\subsection{Accuracy and Generalizability}

Our transformer-based stem-and-trunk design integrated multiview and volumetric inputs, synthesizing information more efficiently than traditional CNN feature concatenation. While CNNs limit cross-view interactions with shared kernels, our transformer architecture utilized self-attention to enhance the learning of cross-view relationships. Although 3D-transformer models showed slightly better MAE under certain conditions, they were less reliable for age estimation in older individuals. Our model benefited from heterogeneous pretraining on robotic proprioception and vision datasets, improving its robustness across varied environments. This approach led to stable performance, as evidenced by consistent accuracy on the AIBL dataset despite domain shifts.

\subsection{Categorical Interpretability}
In Sec. \ref{subsec:result_disorder}, we found that the severity of brain disorder correlated positively with increase in BAG. The mean BAG values increased notably across different disorder groups: 0.15 years in CN, 2.55 years in MCI, and 6.12 years in AD group. This trend aligned with findings from prior studies \cite{bashyam2020mri, he2021global, zhao2023modeling}, supporting the idea that greater cognitive decline correlates with larger deviations from normal brain aging. Bashyam et al. \cite{bashyam2020mri} emphasized that high accuracy within the CN group alone is insufficient for evaluating the effectiveness of a brain age prediction model. Instead, they emphasized the need to measure the contrast and discrepancy in BAG or MAE across diverse disorder groups, as models excelling in CN populations may overlook regional differences caused by subtle pathological changes. Our model demonstrated competitive performance, with BAG contrast between CN and both MCI and AD groups comparable to findings in Lee et al. \cite{lee2022deep} and Zhao et al. \cite{zhao2023modeling} and showed stronger group separability than previous results (MAE of 4.33 years for MCI and 7.05 years for AD, compared to 5.15 and 5.47 years in earlier work \cite{bashyam2020mri}).

\subsection{Functional Interpretability}
In Sec. \ref{subsec:result_disorder}, we explored relationships between BAG and cognitive scores (MMSE and MoCA) and observed significant negative correlations in both the MCI and AD groups, but not in the CN group. The lack of correlation in the CN group aligned with expectations, as cognitive scores typically exhibited minimum variability in this population. The AD group showed a stronger negative correlation than the MCI group, reflecting the more severe cognitive decline presented in AD. These associations have been sparsely explored in the literature \cite{franke2012longitudinal}, making our strong correlation between BAG predictions and cognitive scores significant and interpretable, thus suggesting potential for future studies exploring neuroimaging biomarkers of cognitive decline.

\subsection{Biological Interpretability}
The gradient map from our model predominantly highlighted the white matter, ventricles, caudate nuclei and thalami, consistent with prior cross-sectional and longitudinal studies \cite{bauer2015significance, fama2015thalamic, fujita2023characterization,uchida2024acceleration}. In particular, the ALIC and PLIC were listed among the most critical regions for age prediction, congruent with a prior study \cite{giorgio2010age} reported a strong linear decline in ALIC and PLIC volumes with age. The ALIC contains the frontopontine fibers involved in motor planning and coordination \cite{emos2023neuroanatomy}, the thalamocortical fibers that contribute to cognitive and executive functions \cite{axer1999morphological, george2019neuroanatomy}, and corticostriatal and caudatoputaminal fibers involved in the reward system and motor modulation \cite{mithani2020anterior}. The PLIC contains the corticospinal and corticobulbar tracts essential for motor functions \cite{emos2023neuroanatomy}, and the superior thalamic, optic, and auditory radiations responsible for transmitting sensory information \cite{ramos2019supratentorial}. Research showed a significant linear negative correlation between age and the volume of these fiber tracts \cite{giorgio2010age, kennedy2009pattern}. The decline in the volume and integrity of the ALIC and PLIC is primarily linked to degeneration in myelinated fibers due to aging \cite{gong2014aging, sullivan2010fiber}. Our model captured these biological changes, offering potential utility for early pathological assessments.


\subsection{Limitations and Future Work}

Our current work is based on a cross-sectional study, and tracking individual brain age trajectories over time may offer deeper insights into the relationship between BAG and AD. In addition, we observed a decelerated trend of BAG in older individuals in the AD group, contradicting the common expectation that AD typically accelerates brain aging \cite{lee2022deep}. It is likely due to the under-representation of subjects over 85 years of age in the training dataset and the model predicts the mean.

\subsection{Conclusion}
\label{sec:conclusion}

We introduced the interpretable OpenMAP-BrainAge, a model that effectively combines different views and volumetric data using a stem-and-trunk architecture. Our model achieved an MAE of 3.65 years on ADNI2 $\&$ 3 and OASIS 3, and 3.54 years on the external AIBL dataset. It also improved computational efficiency from quadratic to linear for better scalability with 3D MRI data. We found a positive association between BAG and neurodegenerative disorders along with cognitive decline. Future work with longitudinal data and larger cohorts may enhance insights into brain aging and aid in the detection of neurodegenerative disorders.
\section*{Tables}
\begin{table}[H]
\begin{threeparttable}
\centering
\setlength{\tabcolsep}{10pt}%
 \resizebox{18cm}{!}{
\begin{tabular}{@{}lccc@{}}
\toprule
\textbf{Models} & \textbf{Input} & \textbf{Test MAE (year) ↓} & \textbf{R-value ↑} \\
\midrule
MLP & Volume & 11.43 & 0.33 \\
3D ResNet18  & Pseudo 3D - Coronal & 3.78 & 0.81 \\
3D ResNet18  & Pseudo 3D - three-view + Volume & 4.05 & 0.78 \\
3D ResNet18& Full 3D & 3.61 & 0.82 \\
3D ViT & Pseudo 3D - Coronal & 3.63 & 0.81 \\
3D ViT  & Full 3D & Not applicable & Not applicable \\
SFCN & Full 3D & 5.94 & 0.58 \\
Global-Local Transformer  & Full 3D (average through 2D Axial slices) & 3.87 & 0.74 \\
\midrule
\textbf{OpenMAP-BrainAge} & \textbf{Pseudo 3D - three-view + Volume} & \textbf{3.65} & \textbf{0.82} \\
\bottomrule
\end{tabular}
}
\caption{Model comparison on brain age prediction in CN group of the test split of ADNI 2\&3 \cite{petersen2010alzheimer} and OASIS 3  \cite{lamontagne2019oasis}.} 
\label{tab:result_cn}
\end{threeparttable}
\end{table}

\begin{table}[H]
\centering
\begin{threeparttable}
\setlength{\tabcolsep}{10pt}%
 \resizebox{18cm}{!}{
\begin{tabular}{@{}lccc@{}}
\toprule
\textbf{Models} & \textbf{Input} & \textbf{Test MAE (year) ↓} & \textbf{R-value ↑} \\
\midrule
3D ResNet18  & Pseudo 3D - Coronal & 4.01 & 0.71 \\
3D ResNet18 & Pseudo 3D - three-view + Volume & 3.98 & 0.70 \\
3D ResNet18  & Full 3D & 3.78 & 0.74 \\
3D ViT & Pseudo 3D - Coronal & 3.77 & 0.72 \\
3D ViT  & Full 3D & Not applicable & Not applicable \\
Global-Local Transformer  & Full 3D (average through 2D Axial slices) & 4.48 & 0.64 \\
\midrule
\textbf{OpenMAP-BrainAge} & \textbf{Pseudo 3D - three-view + Volume} & \textbf{3.54} & \textbf{0.76} \\
\bottomrule
\end{tabular}
}
\caption{Model comparison on brain age prediction in CN group of the AIBL dataset \cite{ellis2009australian}. AIBL dataset is demographically different from the datasets during training. Our proposed model has a consistent performance under domain shifting.}
\label{tab:result_aibl}
\end{threeparttable}
\end{table}

\section*{Figures}

\newpage

\section*{Data and Code Availability}

The source code and implementation details for our experiments are publicly available at: \href{https://github.com/pkan2/OpenMAP-BrainAge}{\url{https://github.com/pkan2/OpenMAP-BrainAge}}.

The OpenMAP-T1 parcellation tool used in this study is available at: \href{https://github.com/OishiLab/OpenMAP-T1}{\url{https://github.com/OishiLab/OpenMAP-T1}}.

Data used in the preparation of this article were obtained from the Alzheimer's Disease Neuroimaging Initiative (ADNI) database (adni.loni.usc.edu). The ADNI was launched in 2003 as a public-private partnership, led by Principal Investigator Michael W. Weiner, MD. The original goal of ADNI was to test whether serial magnetic resonance imaging (MRI), positron emission tomography (PET), other biological markers, and clinical and neuropsychological assessment can be combined to measure the progression of mild cognitive impairment (MCI) and early Alzheimer's disease (AD). The current goals include validating biomarkers for clinical trials, improving the generalizability of ADNI data by increasing diversity in the participant cohort, and to provide data concerning the diagnosis and progression of Alzheimer's disease to the scientific community. For up-to-date information, see \href{www.adni.loni.usc.edu}{adni.loni.usc.edu}.

Data used in the preparation of this article was obtained from the Australian Imaging Biomarkers and Lifestyle flagship study of ageing (AIBL). See \href{www.aibl.csiro.au}{www.aibl.csiro.au} for further details.

Open Access Series of Imaging Studies 3 (OASIS 3) dataset is accessed at \href{https://sites.wustl.edu/oasisbrains/home/oasis-3/}{\url{https://sites.wustl.edu/oasisbrains/home/oasis-3/}}.

\section*{Acknowledgements}
Data collection and sharing for this project was funded by the Alzheimer's Disease Neuroimaging Initiative
(ADNI) (National Institutes of Health Grant U01 AG024904) and DOD ADNI (Department of Defense award
number W81XWH-12-2-0012). ADNI is funded by the National Institute on Aging, the National Institute of
Biomedical Imaging and Bioengineering, and through generous contributions from the following: AbbVie,
Alzheimer’s Association; Alzheimer’s Drug Discovery Foundation; Araclon Biotech; BioClinica, Inc.; Biogen;
Bristol-Myers Squibb Company; CereSpir, Inc.; Cogstate; Eisai Inc.; Elan Pharmaceuticals, Inc.; Eli Lilly and
Company; EuroImmun; F. Hoffmann-La Roche Ltd and its affiliated company Genentech, Inc.; Fujirebio; GE
Healthcare; IXICO Ltd.; Janssen Alzheimer Immunotherapy Research \& Development, LLC.; Johnson \&
Johnson Pharmaceutical Research \& Development LLC.; Lumosity; Lundbeck; Merck \& Co., Inc.; Meso
Scale Diagnostics, LLC.; NeuroRx Research; Neurotrack Technologies; Novartis Pharmaceuticals
Corporation; Pfizer Inc.; Piramal Imaging; Servier; Takeda Pharmaceutical Company; and Transition
Therapeutics. The Canadian Institutes of Health Research is providing funds to support ADNI clinical sites
in Canada. Private sector contributions are facilitated by the Foundation for the National Institutes of Health
(www.fnih.org). The grantee organization is the Northern California Institute for Research and Education,
and the study is coordinated by the Alzheimer’s Therapeutic Research Institute at the University of Southern
California. ADNI data are disseminated by the Laboratory for Neuro Imaging at the University of Southern
California.

Data collection in Open Access Series of Imaging Studies 3 is provided through OASIS-3: Longitudinal Multimodal Neuroimaging: Principal Investigators: T. Benzinger, D. Marcus, J. Morris; NIH P30 AG066444, P50 AG00561, P30 NS09857781, P01 AG026276, P01 AG003991, R01 AG043434, UL1 TR000448, R01 EB009352. AV-45 doses were provided by Avid Radiopharmaceuticals, a wholly owned subsidiary of Eli Lilly.

{
\bibliographystyle{ieee_fullname}
\bibliography{sample}
}

\newpage

\appendix
\renewcommand{\thefigure}{S\arabic{figure}}
\setcounter{figure}{0}
\section{Transformer and Attention Mechanism}
\label{ap:attn}
Vision Transformers (ViTs) \cite{dosovitskiy2020image} partition an input image into a sequence of fixed-size patches, referred to as tokens, and model their relationships using the self-attention mechanism \cite{vaswani2017attention}. This mechanism enables the network to integrate information globally by learning pairwise interactions between tokens. The attention mechanism operates on three components: keys $K \in \mathbb{R}^{N_k \times d_k}$, values $V\in \mathbb{R}^{N_k \times d_v}$ and queries $Q\in \mathbb{R}^{N_q \times d_k}$. Keys and values represent the contextual information from the input tokens, while queries extract relevant features by attending to this context. The output of attention mechanism is calculated as:

\begin{align}
    \text{attn}(K, V, Q) = \text{softmax}(\frac{QK^T}{\sqrt{d_k}})V
\end{align}
Each query attends to all key–value pairs, resulting in time and space complexity of $O(N_q \cdot N_k)$. For vision tasks, where $N_q = N_k = n$ (number of image patches), this results in quadratic complexity $O(n^2)$, which limit the scalability in high-resolution images or 3D volumes.

\section{Computational Complexity of Stem}
\label{ap:complexity_stem}
The stem module \cite{wang2024scaling} compresses information by applying attention from a fixed number of $m$ learnable queries  to the encoder outputs of $n$ tokens, where $m \ll n$. This formulation reduces the attention computation from $O(n^2)$ to $O(m \cdot n)$, which scales linearly with the number of input tokens. Since $m$ is fixed and independent of input size, the overall time and space complexity becomes $O(n)$, enabling efficient handling of high-dimensional medical images such as 3D MRIs.
\section{Interpretation with Gradient Mapping}
\label{ap:grad_map}

To interpret our proposed model, we consider gradient-based feature attribution to highlight the critical regions influenced by brain aging. Specifically, for each scanning input $i$, the spatial information contributes to the model's prediction $\hat{y}_i$ through its three anatomical views, denoted as $s_i, c_i$ and $a_i$. We calculate the gradients of these three anatomical views with respect to the training loss $L_\text{training}(\hat{y}_i, y_i)$ and get:

\begin{align}
    g_{s_i}&=\frac{\partial L_\text{training}(\hat{y}_i, y_i)}{\partial s_i}\\
    g_{c_i}&=\frac{\partial L_\text{training}(\hat{y}_i, y_i)}{\partial c_i}\\
    g_{a_i}&=\frac{\partial L_\text{training}(\hat{y}_i, y_i)}{\partial a_i}
\end{align}

It is remarked that these gradients contain sign information. To count for and balance off the effect that each pixel may contribute differently to the final prediction through different anatomical views, we will take the signed addition of the gradients from each view and then take the absolute value to reconstruct the full 3D gradient for the scanning input $i$:

\begin{align}
    G_i &= abs( g_{s_i} +  g_{c_i} +  g_{a_i})
\end{align}

 To offset the spatial difference of each inputs, we calculate the nonlinear transformation $T_i$ to register each input $i$ to the Johns Hopkins University Atlas template \cite{wu2016resource} and use this transformation to register the gradient mapping of each individuals. To generalize the interpretation, we take the average of the gradient mapping across the CN group of the test split:
 
 \begin{align}
     \Bar{G} &=  \frac{1}{N}\sum\limits_{i=1}^N T_i(G_i)
     \label{eq:avg_grad_mapping}
 \end{align}
\section{Supplemental Figures}

\begin{figure}[htbp]
    \centering
    \includegraphics[width=0.8\linewidth]{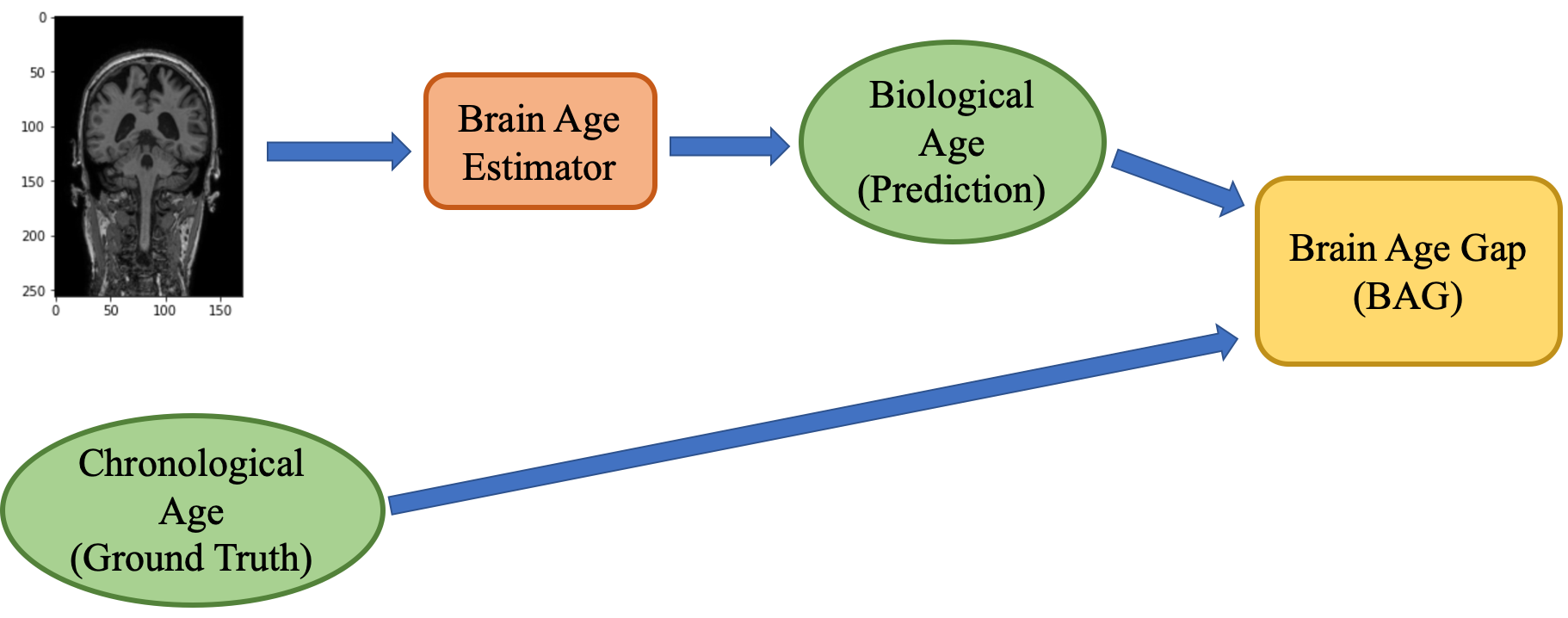}
    \caption{General illustration of brain age estimation and brain age gap (BAG), based on a trained estimator. The brain age estimator is trained over cognitive normal (CN) subjects.}
    \label{fig:age_prediction}
\end{figure}

\begin{figure}[htbp]
    \centering
    \includegraphics[width=1\linewidth]{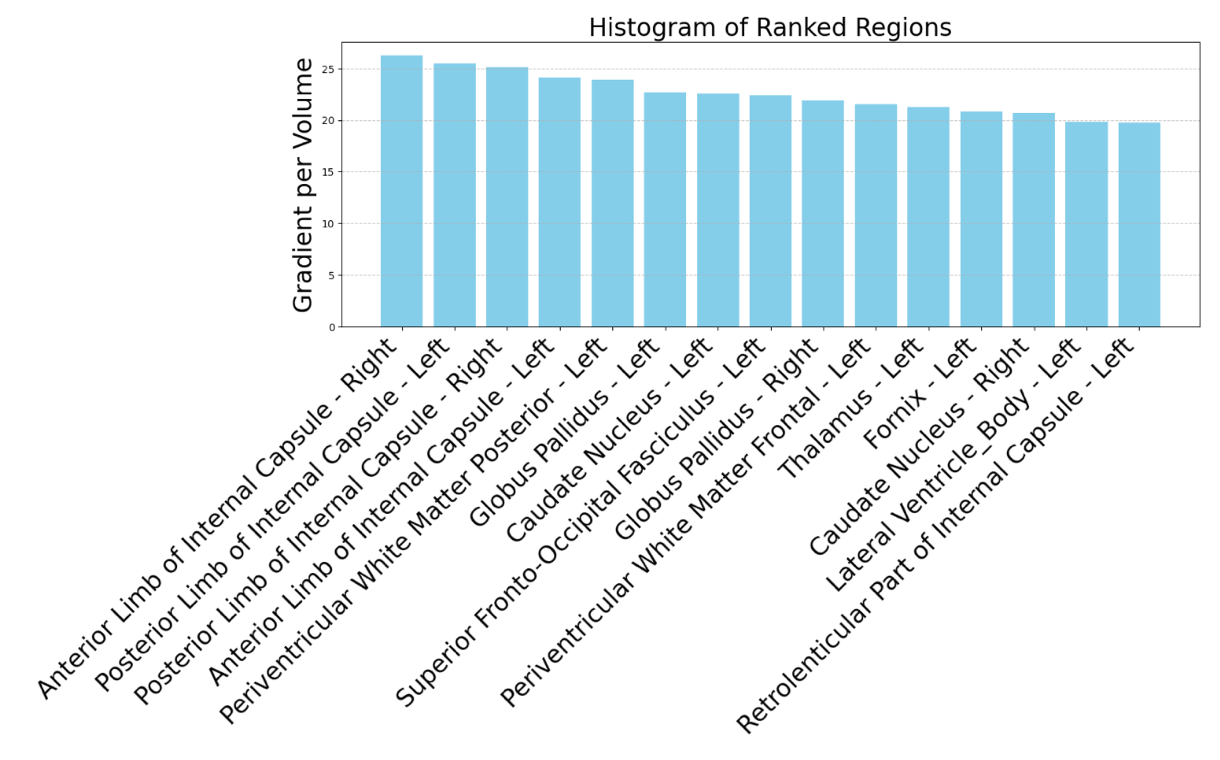}
    \caption{Top 15 Ranking of critical regions influenced by brain aging based on gradient per volume.}
    \label{fig:region_ranking}
\end{figure}

\begin{figure}[h]
    \centering
    \captionsetup[subfigure]{font=footnotesize}
    \begin{subfigure}[b]{0.35\textwidth}
        \includegraphics[width=0.85\textwidth]{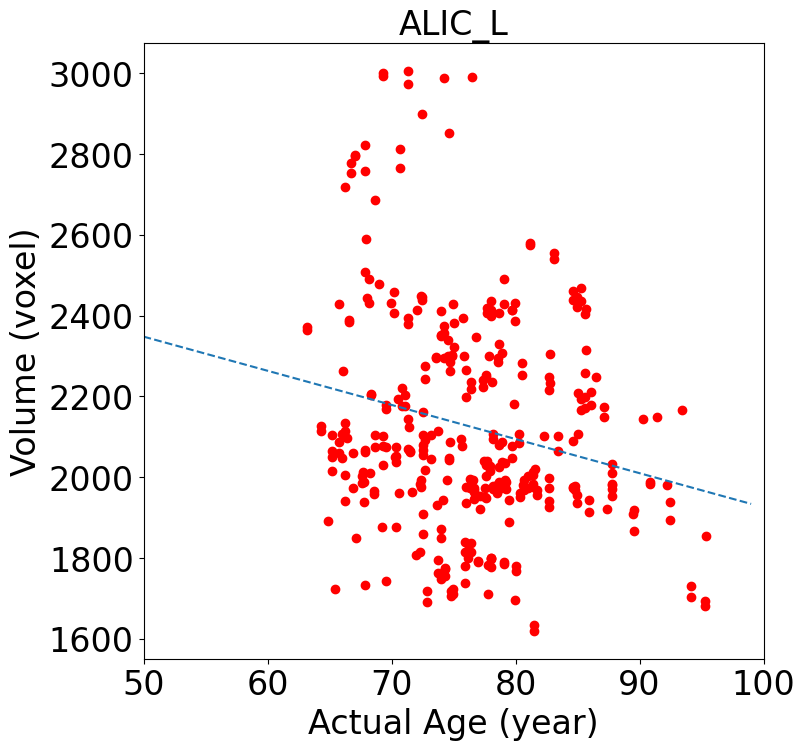}
        \caption{\centering R-value: $-0.21$ (P-value $< 0.001$)}
        \label{fig:vol_a}
    \end{subfigure}
    \begin{subfigure}[b]{0.35\textwidth}
        \includegraphics[width=0.85\textwidth]{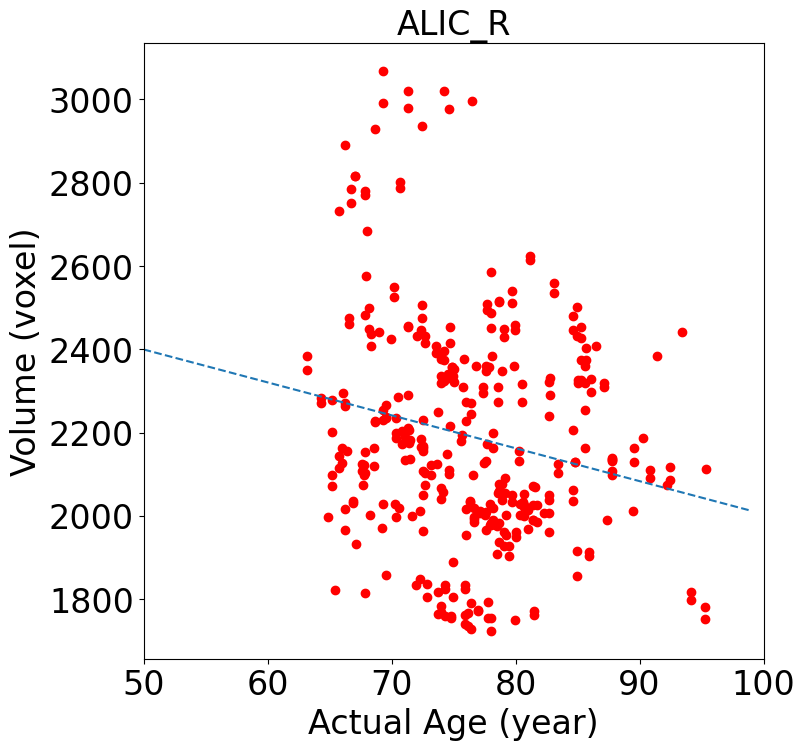}
         \caption{\centering R-value: $-0.20$ (P-value $< 0.001$)}
         \label{fig:vol_b}
    \end{subfigure}
    \hfill
    \begin{subfigure}[b]{0.35\textwidth}
        \includegraphics[width=0.85\textwidth]{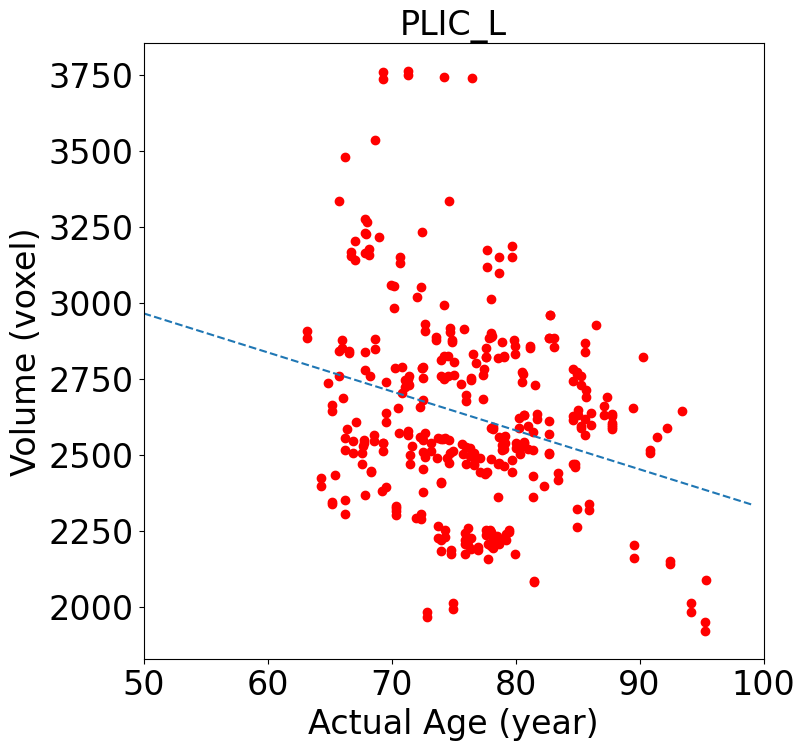}
        \caption{\centering R-value: $-0.27$ (P-value $< 0.001$)}
         \label{fig:vol_c}
    \end{subfigure}
    \begin{subfigure}[b]{0.35\textwidth}
        \includegraphics[width=0.85\textwidth]{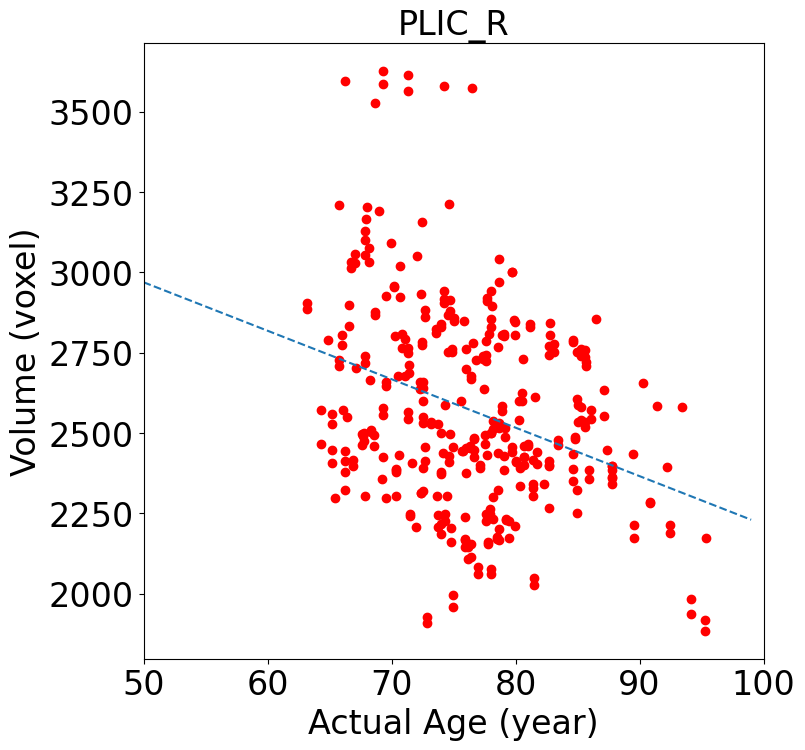}
        \caption{\centering R-value: $-0.33$ (P-value $< 0.001$)}
         \label{fig:vol_d}
    \end{subfigure}
    \hfill 
   \begin{subfigure}[b]{0.35\textwidth}
        \includegraphics[width=0.85\textwidth]{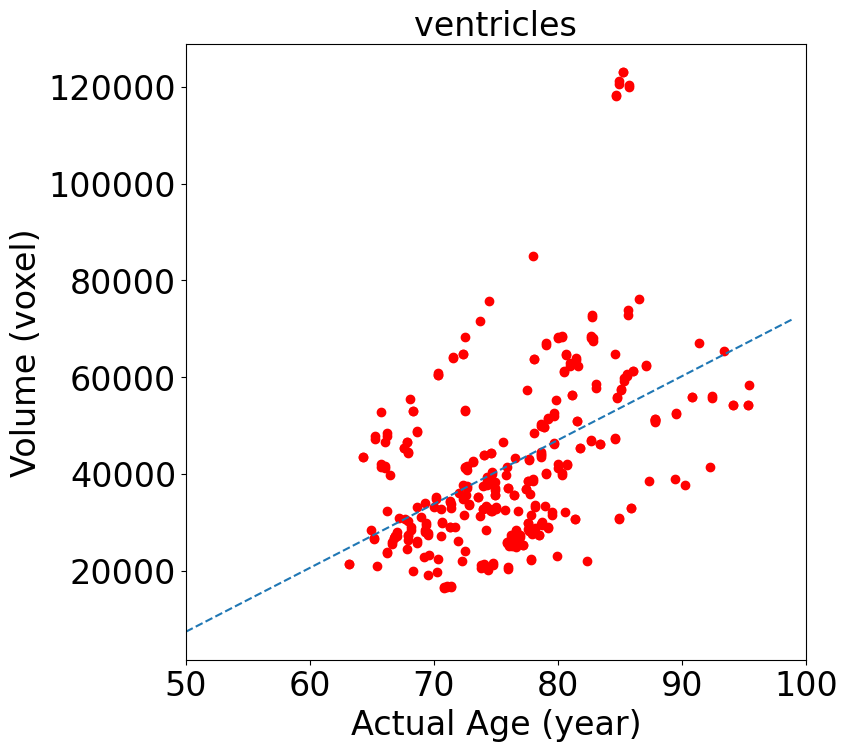}
        \caption{\centering R-value: $0.48$ (P-value $< 0.001$)}
         \label{fig:vol_g}
    \end{subfigure}
    \caption{Correlation analysis between local volumes and age. We observed significant negative correlations between age and the volumes of the Anterior Limb of the Internal Capsule (ALIC) and Posterior Limb of the Internal Capsule (PLIC). In addition, ventricular regions exhibited a trend of volume enlargement with increasing age.}
    \label{fig:vol_age} 
\end{figure}

\end{document}